\documentclass[11pt]{article}
\usepackage{times}
\usepackage{graphicx}
\usepackage[psamsfonts]{amssymb}
\usepackage{amsbsy}

\makeatletter
\def\maketitle{\par 
\begingroup
   \def\thefootnote{\fnsymbol{footnote}}
   \def\@makefnmark{\hbox to 0pt{$^{\@thefnmark}$\hss}} 
   \long\def\@makefntext##1{\parindent 1em\noindent \hbox to1.8em{\hss $\m@th ^{\@thefnmark}$}##1}
   \@maketitle \@thanks
\endgroup
\setcounter{footnote}{0}
\let\maketitle\relax \let\@maketitle\relax
\gdef\@thanks{}\gdef\@author{}\gdef\@title{}\let\thanks\relax}
\makeatother

\makeatletter
\def\@maketitle{\vbox{\hsize\textwidth
\linewidth\hsize \vskip 0.1in \toptitlebar \centering
{\LARGE\bf \@title\par}  \bottomtitlebar 
   \def\And{\end{tabular}\hfil\linebreak[0]\hfil
            \begin{tabular}[t]{c}\bf\rule{\z@}{24pt}\ignorespaces}%
   \def\AND{\end{tabular}\hfil\linebreak[4]\hfil
            \begin{tabular}[t]{c}\bf\rule{\z@}{24pt}\ignorespaces}%
   \def\LINEBREAK{\end{tabular}\linebreak[4]\begin{tabular}[t]{c}\bf\rule{\z@}{16pt}\ignorespaces}%
    \begin{tabular}[t]{c}\bf\rule{\z@}{24pt}\@author\end{tabular}%
\vskip 0.3in minus 0.1in}}
\makeatother

\renewenvironment{abstract}{\vskip.075in\centerline{\large\bf Abstract}\vspace{0.5ex}\begin{quote}}{\par\end{quote}\vskip 1ex}

\makeatletter
\def\section{\@startsection {section}{1}{\z@}{-2.0ex plus -0.5ex minus -.2ex}{1.5ex plus 0.3ex minus0.2ex}{\large\bf\raggedright}}
\def\subsection{\@startsection{subsection}{2}{\z@}{-1.8ex plus-0.5ex minus -.2ex}{0.8ex plus .2ex}{\normalsize\bf\raggedright}}
\def\subsubsection{\@startsection{subsubsection}{3}{\z@}{-1.5ex plus -0.5ex minus -.2ex}{0.5ex plus .2ex}{\normalsize\bf\raggedright}}
\def\paragraph{\@startsection{paragraph}{4}{\z@}{1.5ex plus 0.5ex minus .2ex}{-1em}{\normalsize\bf}}
\def\subparagraph{\@startsection{subparagraph}{5}{\z@}{1.5ex plus  0.5ex minus .2ex}{-1em}{\normalsize\bf}}

\makeatother

\skip\footins 9pt plus 4pt minus 2pt
\def\footnoterule{\kern-3pt \hrule width 12pc \kern 2.6pt }
\leftmargini\leftmargin \leftmarginii 2em
\makeatletter
\def\@listi{\leftmargin\leftmargini}
\def\@listii{\leftmargin\leftmarginii
   \labelwidth\leftmarginii\advance\labelwidth-\labelsep
   \topsep 2pt plus 1pt minus 0.5pt
   \parsep 1pt plus 0.5pt minus 0.5pt
   \itemsep \parsep}
\def\@listiii{\leftmargin\leftmarginiii
    \labelwidth\leftmarginiii\advance\labelwidth-\labelsep
    \topsep 1pt plus 0.5pt minus 0.5pt 
    \parsep \z@ \partopsep 0.5pt plus 0pt minus 0.5pt
    \itemsep \topsep}
\def\@listiv{\leftmargin\leftmarginiv
     \labelwidth\leftmarginiv\advance\labelwidth-\labelsep}
\def\@listv{\leftmargin\leftmarginv
     \labelwidth\leftmarginv\advance\labelwidth-\labelsep}
\def\@listvi{\leftmargin\leftmarginvi
     \labelwidth\leftmarginvi\advance\labelwidth-\labelsep}
\makeatother

\belowdisplayskip \abovedisplayskip
\makeatletter
\def\normalsize{\@setsize\normalsize{11pt}\xpt\@xpt}
\def\small{\@setsize\small{10pt}\ixpt\@ixpt}
\def\footnotesize{\@setsize\footnotesize{10pt}\ixpt\@ixpt}
\def\scriptsize{\@setsize\scriptsize{8pt}\viipt\@viipt}
\def\tiny{\@setsize\tiny{7pt}\vipt\@vipt}
\def\large{\@setsize\large{14pt}\xiipt\@xiipt}
\def\Large{\@setsize\Large{16pt}\xivpt\@xivpt}
\def\LARGE{\@setsize\LARGE{20pt}\xviipt\@xviipt}
\def\huge{\@setsize\huge{23pt}\xxpt\@xxpt}
\def\Huge{\@setsize\Huge{28pt}\xxvpt\@xxvpt}
\makeatother

\def\toptitlebar{\hrule height4pt\vskip .25in\vskip-\parskip}
\def\bottomtitlebar{\vskip .29in\vskip-\parskip\hrule height1pt\vskip .09in}

\title{Modeling Image Structure with Factorized Phase-Coupled Boltzmann Machines}

\author{Charles F.~Cadieu \hspace{1ex}\&\hspace{1ex} Kilian Koepsell\\[2ex]
Redwood Center for Theoretical Neuroscience \\
Helen Wills Neuroscience Institute \\
University of California, Berkeley \\
Berkeley, CA 94720 \\
\texttt{\{cadieu, kilian\}@berkeley.edu}
}

\begin{document}

\maketitle

\begin{abstract}
We describe a model for capturing the statistical structure of local amplitude
and local spatial phase in natural images. The model is based on a recently
developed, factorized third-order Boltzmann machine that was shown to be
effective at capturing higher-order structure in images by modeling
dependencies among squared filter outputs~\cite{ranzato2010}. Here, we extend
this model to $L_p$-spherically symmetric subspaces.  In order to model local
amplitude and phase structure in images, we focus on the case of two
dimensional subspaces, and the $L_2$-norm. When trained on natural images the
model learns subspaces resembling quadrature-pair Gabor filters. We then
introduce an additional set of hidden units that model the dependencies among
subspace phases. These hidden units form a combinatorial mixture of phase
coupling distributions, concentrated in the sum and difference of phase
pairs. When adapted to natural images, these distributions capture local
spatial phase structure in natural images.
\end{abstract}

\section{Introduction}
In recent years a number of models have emerged for describing higher-order
structure in images (i.e., beyond sparse, Gabor-like decompositions).  These
models utilize distributed representations of covariance matrices to form an
infinite or combinatorial mixture of Gaussians model of the
data~\cite{karklin2008emergence, ranzato2010}. These models have been shown to
effectively capture the non-stationary variance structure of natural images. A
variety of related models have focused on the local radial (in vectorized
image space) structure of natural
images~\cite{Lyu2009,Lyu2009a,Sinz2008,Koster2007,CadieuNIPS2009}. While these
models represent a significant step forwards in modeling higher-order natural
image structure, they only implicitly model local phase alignments across
space and scale. Such local phase alignments are implicated as being a
hallmark of edges, contours, and other shape structure in natural
images~\cite{kovesi2000}. The model proposed in this paper attempts to extend
these models to capture both amplitude and phase in natural images.

In this paper, we first extend the recent factorized, third-order Boltzmann
machine model of Ranzato \& Hinton~\cite{ranzato2010} to the case of
$L_p$-spherically symmetric distributions.  In order to directly model the
dependencies among local amplitude and phase variables, we consider the
restricted case of two-dimensional subspaces with $L_2$-norm. When adapted to
natural images, the subspace filters converge to quadrature-pair Gabor-like
functions, similar to previous work~\cite{CadieuNIPS2009}. The dependencies
among amplitudes are modeled using a set of hidden units, similar to Ranzato
\& Hinton~\cite{ranzato2010}.  Phase dependencies between subspaces are
modeled using another set of hidden units as a mixture of phase coupling
``covariance'' matrices: conditioned on the hidden units, phases are modeled
via a phase-coupled distribution~\cite{cadieu2010}.

\subsection{Modeling Local Amplitude and Phase}
Our model may be viewed within the same framework as a number of recent models
that attempt to capture higher-order structure in images by factorizing the
coefficients of oriented, bandpass
filters~\cite{Lyu2009,Lyu2009a,Sinz2008,Koster2007,CadieuNIPS2009,Karklin2005,Schwartz2001}. These
models are currently the best probabilistic models of natural image structure:
they produce state-of-the-art denoising~\cite{Lyu2009}, and achieve lower
entropy encodings of natural images~\cite{Sinz2008}. They can be viewed as
sharing a common mathematical form in which the filter coefficients, $x$, are
factored into a non-negative component $z$ and a scalar component $u$, where
$x=z\, u$. The non-negative factors, $z$, are modeled as either an independent
set of variables each shared by a pair of linear
components~\cite{CadieuNIPS2009}, a set of variables with learned dependencies
to the linear components~\cite{Koster2007}, or as a single radial
component~\cite{Sinz2008,Lyu2009a}. The scalar factor, $u$, is modeled in a
number of ways: as an independent angular unit
vector~\cite{Sinz2008,Lyu2009a}, as a correlated noise process~\cite{Lyu2009},
as the phase angle of paired filters~\cite{CadieuNIPS2009}, or as a sparse
decomposition of latent variables~\cite{Karklin2005}.

By separating the filter coefficients into two sets of variables, it is
possible for higher levels of analysis to model higher-order statistical
structure that was previously entangled in the filter coefficients themselves.
For example, the non-negative variables $z$ are usually related to the local
contrast or power within an image region, and Karklin \& Lewicki have shown
that it is possible to train a second layer to learn the structure in these
variables via sparse coding~\cite{Karklin2005}. Similarly, Lyu \& Simoncelli
learn an MRF model on these variables and show that the resulting model
achieves state-of-the-art denoising~\cite{Lyu2009}. It is generally less clear
what structure is represented in the scalar variables $u$. Here we take
inspiration from Zetzsche's observations regarding the circular symmetry of
joint distributions of related filter coefficients and conjecture that this
quantity should be modeled as the sine or cosine of an underlying phase
variable, i.e., $u=\sin(\theta)$ or $u=\cos(\theta)$~\cite{Zetzsche1999}.

One of the main contributions of this paper is a model of the joint structure
of local phase in natural images. For the case of phases in a complex pyramid,
the empirical marginal distribution of phases is always uniform across a
corpus of natural images (data not shown); however, the empirical distribution
of pairwise phase differences is often concentrated around a preferred
phase. We can write the joint distribution of a pair of phase variables with a
dependency in the difference of the phase variables as a von Mises\footnote{A
  von Mises distribution for an angular variable $\phi$ with concentration
  parameter, $k$, and mean, $\mu$, is defined as $p(\phi) := \frac{1}{2\pi
    I_0(k)}e^{k\cos(\phi-\mu)}$, where $I_0(k)$ is the zeroth order modified
  Bessel function.} distribution in the difference of the phases:
\begin{equation}
p(\theta_i,\theta_j) = \frac{1}{2\pi I_0(k)}\exp(\kappa\cos(\theta_i-\theta_j-\mu))
\label{eq:vonmises}
\end{equation}
This distribution is parameterized by a concentration $\kappa$ and a phase
offset $\mu$. Using trigonometric identities we can re-express the cosine of
the difference as a sum of bivariate terms of the form
$\cos(\theta_i)\sin(\theta_j)$. One may view these terms as the pairwise
statistics for angular variables, just as the the bivariate terms in the
covariance matrix are the pairwise statistics for a Gaussian. This logic
extends to multivariate distributions with $n>2$.

\begin{figure}
\[
\includegraphics[width=.6\linewidth]{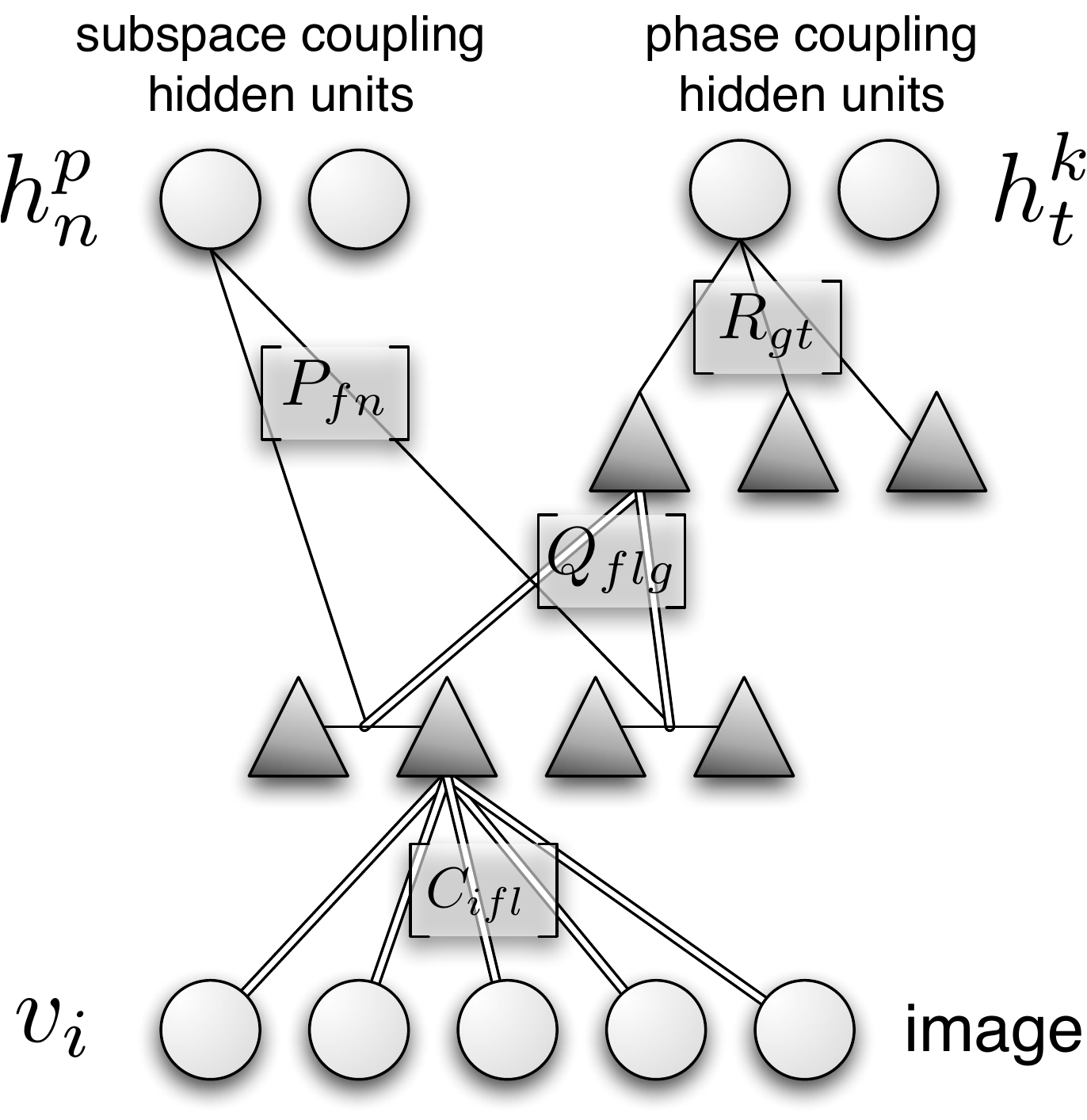}
\]
\caption{The mpkRBM model is shown schematically. The visible units $v_i$ are
  connected to a set of factors through the matrix $C_{ifl}$. These factors
  come in pairs, as indicated by the horizontal line linking neighboring
  triangles. The squared output of each factor connects to the subspace
  coupling hidden units $h_n^p$ through the matrix $P_{fn}$. Another set of
  factors is connected to the sine and cosine representations of each factor
  (second layer triangles) through the matrix $Q_{flg}$. This second set of
  factors connects to the phase coupling hidden units $h_t^k$. Single-stroke
  vertical lines indicate linear interactions, while double stroke lines
  indicate quadratic interactions. Note that the mean factors, $h^m$, and
  biases are omitted for clarity. See text for further explanation.}
\label{fig:model}
\end{figure}

In the next section we describe an extension to the mean and covariance
Restricted Boltzmann Machine (mcRBM) of Ranzato \&
Hinton~\cite{ranzato2010}. Our extension represents local amplitude and phase
in its factors. We model the amplitude dependencies with a set of hidden
variables, and we model the phase dependencies among factors as a
combinatorial mixture of phase-coupling distributions~\cite{cadieu2010}. This
model is shown schematically in Fig.~\ref{fig:model}.

\section{Model}
We first review the factorized third-order Boltzmann machine of Ranzato and
Hinton~\cite{ranzato2010} named mcRBM because it models both the mean and
covariance structure of the data. We then describe an extension that models
pairs of factors as two dimensional subspaces, which we call the mpRBM. The
mpRBM provides a phase angle between pairs. The joint statistics between phase
angles are not explicitly modeled by the mpRBM. Thus we propose additional
hidden factors that model the pair-wise phase dependencies as a product of
phase coupling distributions. For convenience we name this model mpkRBM, where
k references the phase coupling matrix $K$ that is generated by conditioning
on the phase-coupling hidden units.

\subsection{mcRBM}
The mcRBM defines a probability distribution by specifying an energy function
\emph{E}. The probability distribution is then given as $p(v)\propto \exp
(-E(v))$, where $v \in R^D$ is the vectorized input image patch with $D$
pixels. The mcRBM has two major parts of the energy, the covariance
contributions, $E^c$, and the mean contributions, $E^m$.

The covariance terms are defined as:
\begin{equation}
E^c(v,h^c) = -\frac{1}{2}  \sum_{n=1}^N h^c_n \sum_{f=1}^F P_{fn} ( \sum_{i=1}^D C_{if} \frac{v_i}{||v||})^2 - \sum_{n=1}^N b_n^ch_n^c
\end{equation}
where $P \in R^{F\times N}$ is a matrix with positive entries, $N$ is the
number of hidden units and $b^c$ is a vector of biases. The columns of the
matrix $C \in R^{D\times F}$ are the image domain filters and their squared
outputs are weighted by each row in $P$. The $P$ matrix can be considered as a
weighting on the squared outputs of the image filters. We normalize the
visible units ($||v||_{L_2} = 1$) following the procedure
in~\cite{ranzato2010}. Each term in the first sum includes two visible units
and one hidden unit and is a third-order Boltzmann machine. However, the
third-order interactions are restricted by the form of the model into
factors. Each factor, $( \sum_{i=1}^D C_{if} v_i)^2$ is a deterministic
mapping from the image domain. The hidden units combine combinatorially to
produce a zero-mean Gaussian distribution with an inverse covariance matrix
that is a function of the hidden units:
\begin{equation}
\Sigma^{-1} = C\, \mathrm{diag} ( Ph^c ) C'
\label{eq:Sigma}
\end{equation}
Because the representation in the hidden units is distributed, the model can
describe a combinatorial number of covariance matrices.

The mean contribution to the energy, $E^m$ is given as:
\begin{equation}
E^m(v,h^m) = - \sum_{j=1}^M h^m_j \sum_{i=1}^D W_{ij} v_i - \sum_{j=1}^M b_j^m h_j^m
\end{equation}
with $M$ binary hidden units $h_j^m$ that connect directly to the visible
units through the matrix $W_{ij}$. The $b_j^m$ terms are the mean hidden
biases. The form of the mean contribution is a standard RBM
\cite{Hinton2002}. Note that the conditional over both sets of hidden units is
factorial.

The conditional distribution over the visible units given the hidden units is
a Gaussian distribution, which is a function of the hidden variable states:
\begin{equation}
p(v | h^c, h^m) \sim N ( \Sigma (\sum_{j=1}^M W_{ij} h^m_j), \Sigma)
\end{equation}
where $\Sigma$ is given as in Eq.~\ref{eq:Sigma}. The mean of the specified
Gaussian is a function of both the mean $h^m$ and covariance $h^c$ hidden
units.

The total energy is given by:
\begin{equation}
E(v,h^c,h^m) = E^c(v,h^c) + E^m(v,h^m) + \frac{1}{2}\sum_{i=1}^D v_i^2 - \sum_{i=1}^D b_i^v v_i
\end{equation}
with the last two terms a penalty on the variance of the visible units
introduced because $E^c$ is invariant to the norm of the visible units and
biases $b_i^v$ on the visible units.

\subsection{mpRBM}
A number of recent results indicate that the local structure of image patches
is well modeled by $L_p$-spherically symmetric subspaces~\cite{Koster2007}. To
produce $L_p$-spherically symmetric subspaces we impose a pairing of factors
into an $L_p$ subspace. The covariance energy term in the mcRBM is thus
altered to give:
\begin{equation}
E^p(v,h^p) = -\frac{1}{2} \sum_{n=1}^N h^p_n \sum_{f=1}^F P_{fn} \left
[\sum_l^L ( \sum_{i=1}^D C_{ifl} \frac{v_i}{||v||})^\alpha \right ]^{1/\alpha}
- \sum_{n=1}^N b_n^ch_n^c
\end{equation}
Now the tensor $C_{ifl}$ is a set of filters for each factor, $f$ spanning the
$L$ dimensional subspace over the index $l$. The distribution over the visible
conditioned on the hidden units can be expressed as a mixture of $L_p$
distributions. Note that the hidden units remain independent conditioned on
the visible units.

The optimal choice of $L$ and $\alpha$ is an interesting project related to
recent models~\cite{sinz2010} but is beyond the scope of this paper. Here, we
have chosen to focus on modeling the structure in the space complementary to
the norms of the subspaces. To achieve a tractable form of the subspace
structure we select the special case of $L=2$ and $\alpha=2$. The choice of
$\alpha=2$ is motivated by subspace-ICA models~\cite{Koster2007} and sparse
coding with complex basis functions~\cite{CadieuNIPS2009} where the amplitude
within each complex basis function subspace is modeled as a sparse component.

\subsection{mpkRBM}
While the formulation of $L_p$-spherically symmetric subspace models the
spherically symmetric distributions of natural images, there are likely to be
residual dependencies between the subspaces in the non-radial directions. For
example, elongated edge structure will produce dependencies in the phase
alignments across space and through spatial scale~\cite{kovesi2000}. Such
dependencies are not captured, or are at least only implicitly captured in the
mpRBM. By formulating the mpRBM with $L=2$ and $\alpha=2$ we can define a
phase angle within each subspace. The dependencies between these phase angles
will capture image structure such as phase alignments due to edges. We define
a new variable, $x_{fl}$, which is a deterministic function of the visible
units: $x_{f1}=\cos (\theta_f)$ and $x_{f2}=\sin (\theta_f)$, where $\theta_f
= \arg{ \left ( (\sum_{i=1}^D C_{if1} v_i ) +\mathbf{j}( \sum_{i=1}^D C_{if2}
  v_d) \right )}$, and $\mathbf{j}$ is the imaginary unit and $\arg(.)$ is the
complex argument or phase.

We now use a mathematical form that is similar to the covariance model
contribution in the mcRBM to model the joint distribution of phases. We define
the energy of the phase coupling contribution, denoted $E^k$, as,
\begin{equation}
E^k(v,h^k) = -\frac{1}{2} \sum_{t=1}^T h^k_t \sum_{g=1}^G R_{gt}
(\sum_{l=1}^{L}\sum_{f=1}^{F} Q_{flg} x_{fl})^2 - \sum_{t=1}^T b_t^kh_t^k
\end{equation}
with $T$ binary hidden units $h_t^k$ that modulate the columns of the matrix
$R \in R^{G\times T}$. The rows of R then modulate the squared projections of
the vector $x$ through the matrix $Q \in R^{F\times L\times G}$. The term
$b_t^k$ is a vector of biases for the $h_t^k$ hidden units. Similar to the
$h^c$ terms in the mcRBM, the $h_k$ units in the mpkRBM contribute pair-wise
dependencies in the sine-cosine space of the phases. Pair-wise dependencies in
the sine-cosine space can be re-expressed using trigonometric identities as
terms in the sums and differences of the phase pairs, identical to the phase
coupling described in Eq.~\ref{eq:vonmises}. Such explicit dependencies may be
important to model because edges in images exhibit structured dependencies in
the differences of local spatial phase.

Because the phase coupling energy is additive in each $h^k_t$ term the hidden
unit distribution conditioned on the hidden units is factorial. The
probability of a given $h^k_t$ is given as:
\begin{equation}
p(h^k_t | v) = \sigma \left ( \frac{1}{2} \sum_{g=1}^G R_{gt}
(\sum_{l=1}^L\sum_{f=1}^F Q_{flg} x_{fl})^2 + b_t^k \right )
\end{equation}
where the sigmoid, or logistic, function is $\sigma(y) = (1+\exp(-y))^{-1}$.

We can see the dependency structure imposed by the $h^k$ units by considering
the conditional distribution in the space of the phases, $\theta$:
\begin{eqnarray}
K = Q \,\mathrm{diag} ( Rh^k ) Q' \nonumber \\
p(\theta | h^k) \propto \exp( - \scriptstyle \frac{1}{2} \displaystyle x' Kx)
\label{eq:K_coupling}
\end{eqnarray}
Therefore, the $h^k$ units provide a combinatorial code of phase-coupling
dependencies. The number of phase-coupling matrices that the model can
generate is exponential in the number of $h^k$ hidden units because the hidden
unit representation is binary and combinatorial. Again, instead of allowing
arbitrary three way interactions between the $x$ variables and the hidden
units, we have chosen a specific factorization where the squared factors are
$( \sum_{l=1}^{L}\sum_{f=1}^{F} Q_{flg} x_{fl})^2$. Because there are no
direct interactions between the hidden units, $h^k$, the model still has the
form of a conventional Restricted Boltzmann Machine. We call this model a
mpkRBM because it builds upon the mpRBM and the k references the coupling
matrix in the pair-wise phase distribution produced by conditioning on $h^k$.

Combining the three types of hidden units, $h^p$, $h^m$, and $h^k$, allows
each type of hidden unit to model structure captured by the corresponding
functional form. For example, the $h^p$ hidden units will generate phase
dependencies implicitly through their activations. However, if the phase
structure of the data contains additional structure not captured implicitly by
the $h^p$ and $h^m$ hidden units, there will be a learning signal for the
$h^k$ units. Conversely, the phase statistics that are produced implicitly by
the $h^p$ and $h^m$ units will be ignored by the $h^k$ terms because the
learning signal is driven by the differences in the data and model
distributions.

\section{Learning}
We learn the parameters of the model by stochastic gradient ascent of the
log-likelihood. We express the likelihood in terms of the energy with the
hidden units integrated out (omitting the visible squared term and biases):
\begin{eqnarray}
F(v) = &-&\sum_{n=1}^N \log ( 1+ \exp (\frac{1}{2} \sum_{f=1}^F P_{fn}
\left[\sum_l^L ( \sum_{i=1}^D C_{ifl} \frac{v_i}{||v||})^\alpha
  \right]^{1/\alpha} + b_n^c)) \\\nonumber
&-&\sum_{t=1}^T \log ( 1 + \exp (\frac{1}{2} \sum_{g=1}^G R_{gt}
(\sum_{l=1}^L\sum_{f=1}^F Q_{flg} x_{fl})^2 + b_t^k)) \\ \nonumber
&-& \sum_{j=1}^M \log ( 1+ \exp ( \sum_{i=1}^D W_{ij} v_i + b_j^m))
\end{eqnarray}
It is not possible to efficiently sample the distribution over the
visible units conditioned on the hidden units exactly (in contrast, sampling from the
visible units conditioned on the hidden units in a standard RBM is efficient and
exact). We choose to integrate out the hidden variables, instead of 
taking the conditional distribution, to achieve better estimates of the 
model statistics.

Maximizing the log-likelihood the gradient update for the model
parameters (denoted as $\Theta \in \{R, Q, b^k, C, P, b^c, W, b^m, b^v\}$ is
given as:
\begin{equation}
\frac{\partial L}{\partial \Theta} = \langle \frac{\partial F}{\partial
  \Theta} \rangle_{model} - \langle \frac{\partial F}{\partial \Theta}
\rangle_{data}
\end{equation}
where $\langle . \rangle_{\rho}$ indicates the expectation taken over the
distribution $\rho$ . Calculating the expectation over the data distribution
is straightforward. However, calculating the expectation over the model
distribution requires computationally expensive sampling from the equilibrium
distribution. Therefore, we use standard techniques to approximate the
expectation of the gradients under the model distribution following the
procedure in~\cite{hinton2006,ranzato2010}. To summarize, in Contrastive
Divergence learning~\cite{Hinton2002} the model distribution is approximated
by running a dynamic sampler starting at the data for only one step. Given the
energy function with the hidden units integrated out, we run hybrid Monte
Carlo sampling~\cite{Neal1993} starting at the data for one dynamical
simulation to produce an approximate sample from the model distribution. For
each dynamical simulation we draw a random momentum and run 20 leap-frog steps
while adapting the step size to achieve a rejection rate of about ~10\%.
\begin{figure}
\[
\includegraphics[width=\linewidth]{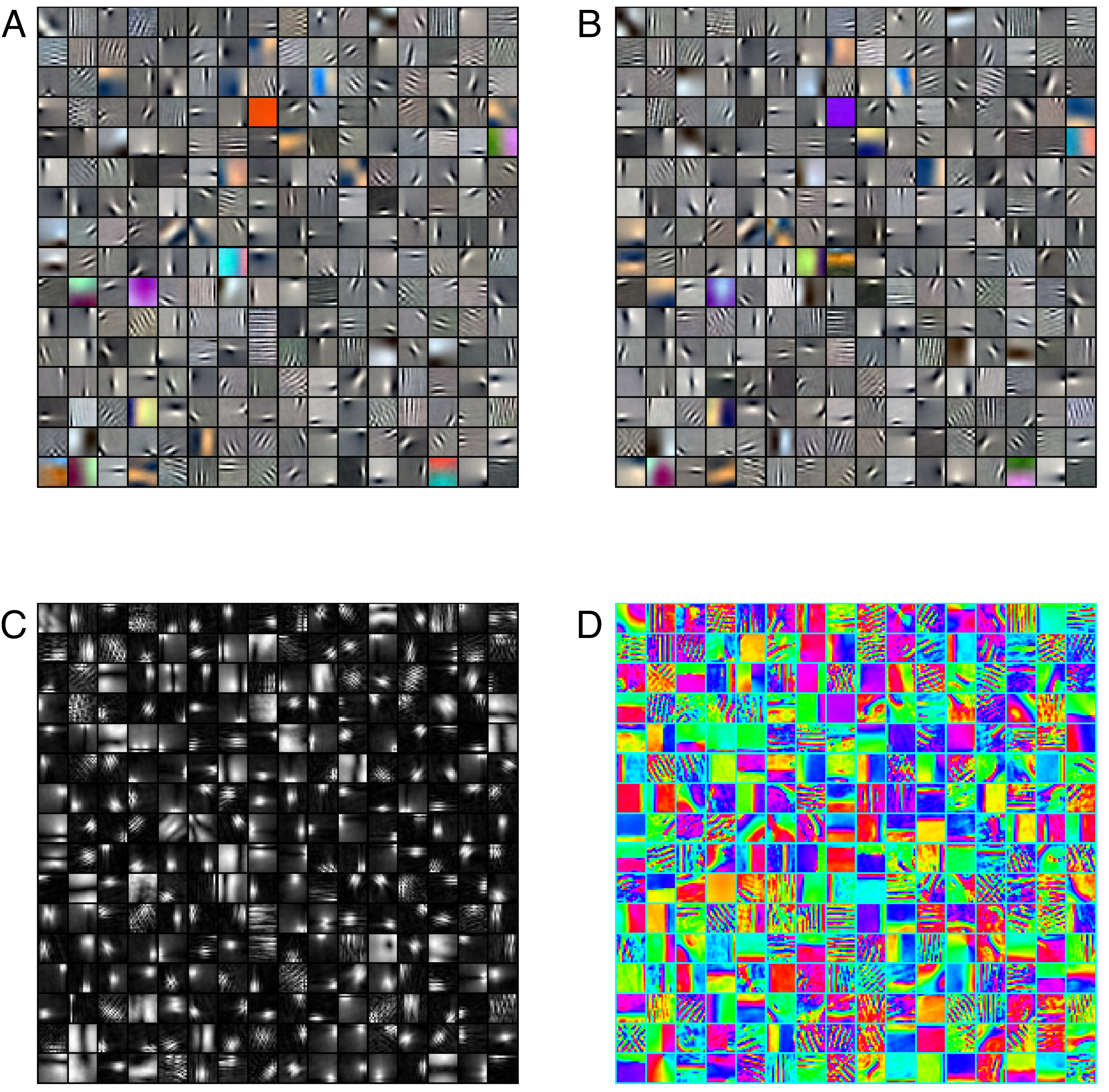}
\]
\caption{Learned $L_p$ Weights, $C$: A) and B) show the individual components
  of each filter pair $C_{if1}$ and $C_{if2}$, respectively. Each subimage
  shows the image domain weights in the unwhitened color image space. C) shows
  the amplitude $\sqrt{(C_{if1}^2+C_{if2}^2)}$ as a function of space, and D)
  shows the phase $\mathrm{arg}(C_{if1}+i C_{if2})$ as a function of
  space. Each panel preserves the ordering such that the image in position
  (1,1) of A) corresponds to the same subspace of $C$ as the image in position
  (1,1) of B), C), and D).}
\label{fig:C_mpkRBM}
\end{figure}

\subsection{Learning parameters}
We trained the models on image patches selected randomly from the Berkeley
Segmentation Database. We subtracted the image mean, and whitened 16x16 color
image patches preserving 99\% of the image variance. This resulted in $D=138$
visible units. We examined a model with 256 $h^c$ covariance units, 256 $h^k$
phase-coupling units, and 100 $h^m$ mean units. We initialized the values of
the matrix $C$ to random values and normalized each image domain vector to
have unit length. We initialized the matrices $W$ and $Q$ to small random
values with variances equal to 0.05, and 0.1 respectively. We initialized the
biases, $b^c$, $b^m$, $b^k$, and $b^v$ to 2.0, -2.0, 0.0, and 0.0
respectively. The learning rates for $R$, $Q$, $P$, $C$, $W$, $bq$, $bc$,
$bm$, $bv$, were set to 0.0015, 0.1, 0.0015, 0.15, 0.015, 0.0005, 0.0015,
0.0075, and 0.0015, respectively.

After each learning update we normalized the lengths of the $C$ vectors to
have the average of the lengths. This allowed the lengths of the $C$ vectors
to grow or shrink to match the data distribution, but prevented any relative
scaling between the subspaces. After each update we also set any positive
values in $P$ to zero and normalized the columns to have unit
$L_2$-norm. Finally, we normalized the lengths of the columns of $R$ to have
unit $L_2$-norm. We learned on mini-batches of 128 image patches and learned
the various parts of the model sequentially. We adapted the parameters of a
mpRBM model with $L=2$ and $\alpha=2$ and fixed the matrix $P$ to the negative
identity for 10,000 iterations. We then adapted the parameters, including $P$,
for another 30,000 iterations. We then added the $h^k$ units to this learned
model and adapted the values in $Q$ for 20,000 iterations while holding the
matrix $R$ fixed to the identity. Next we adapted $R$ for 20,000
iterations. Finally, we allowed all of the parameters in the model to adapt
for 40,000 iterations.

\begin{figure}
\[
\includegraphics[width=.3\linewidth]{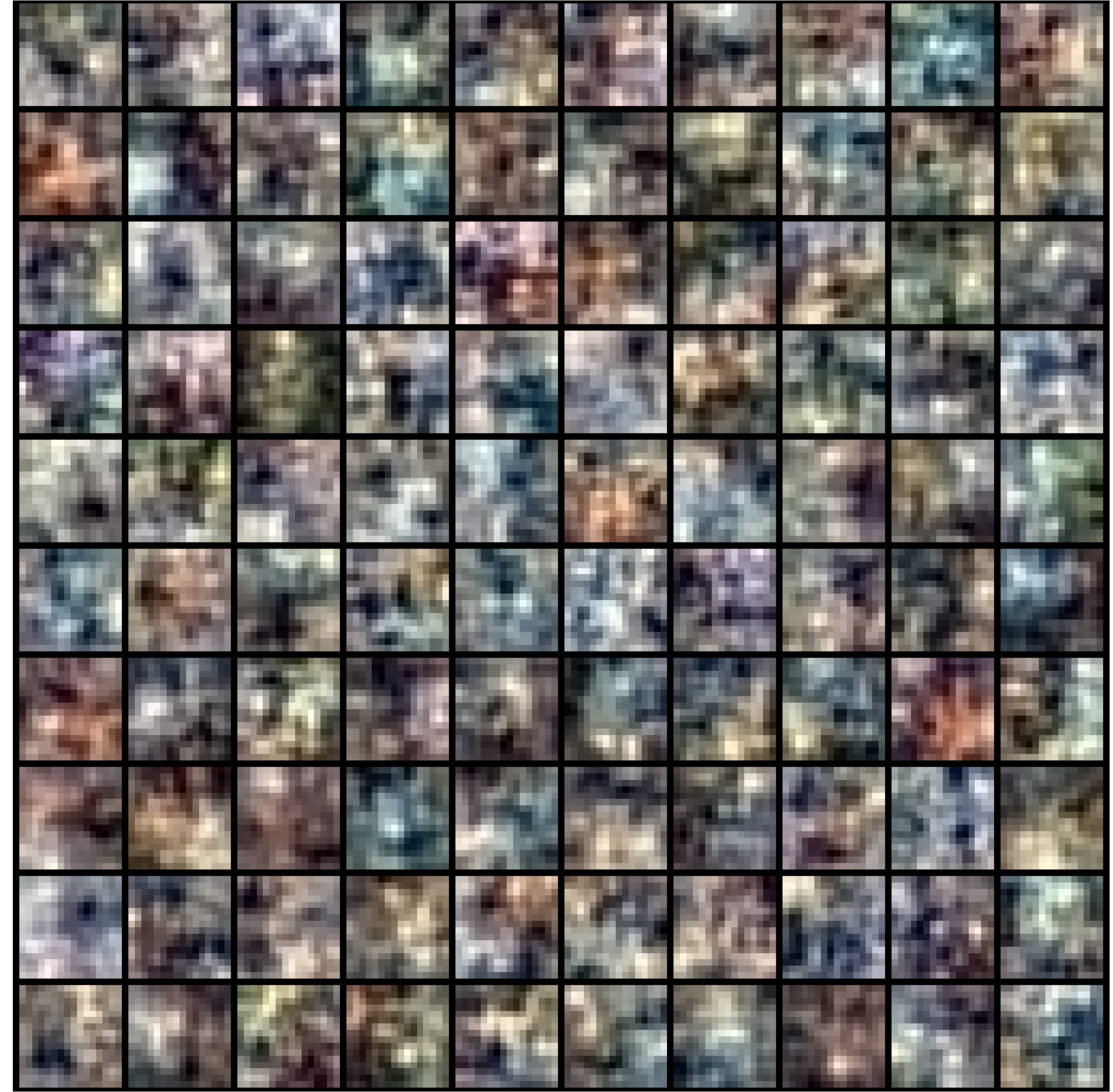}
\]
\caption{Learned Mean Weights, $W$: Each subimage shows a column of $W$ in the
  unwhitened color image domain. These functions resemble those found by the
  mcRBM.}
\label{fig:W}
\end{figure}

\section{Experiments}
\subsection{Learning on Natural Images}
\begin{figure}
\[
\includegraphics[width=\linewidth]{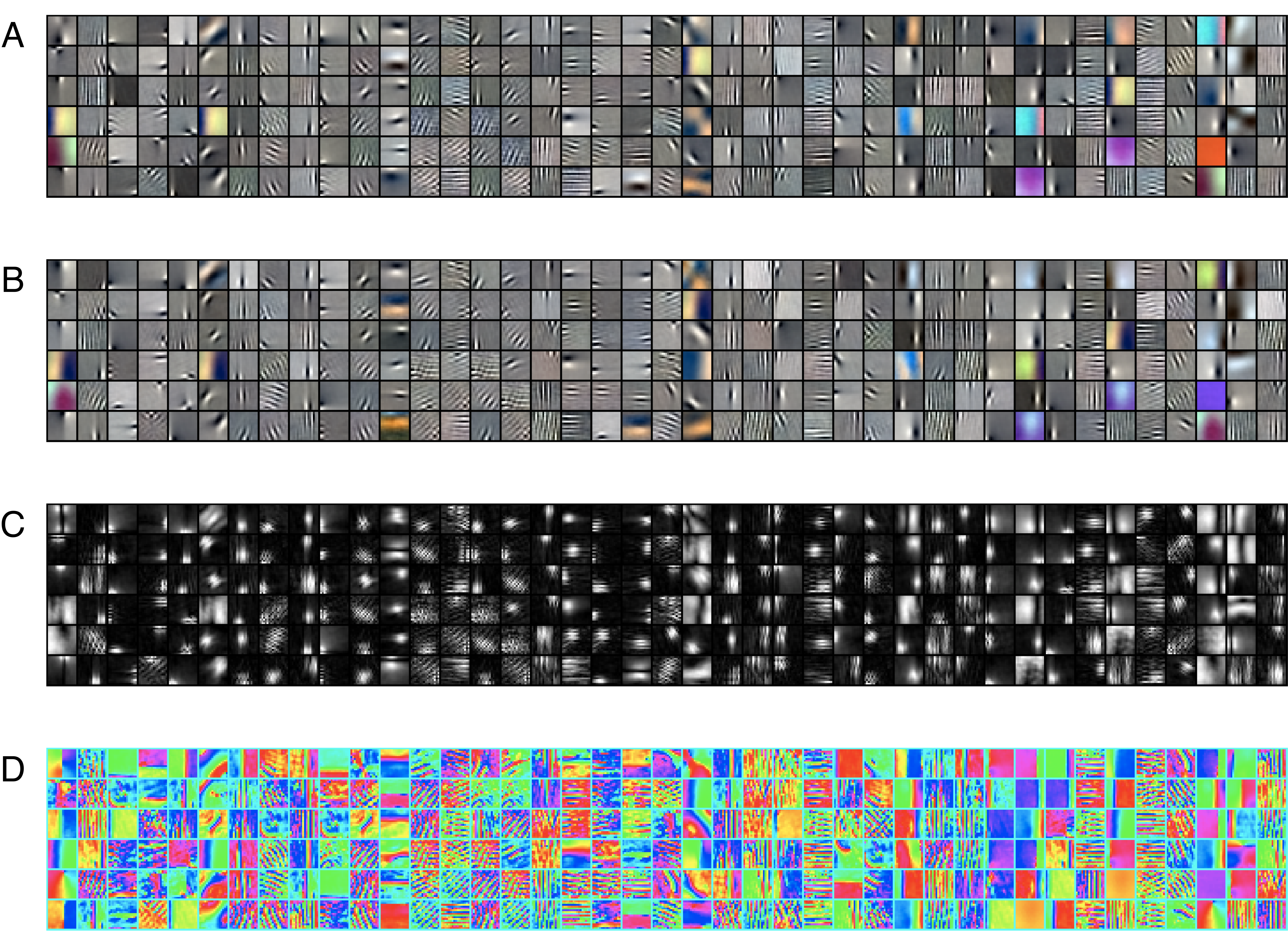}
\]
\caption{Learned $L_p$ groupings, $P$: A random selection (out of 256) columns
  in $P$. Each column depicts the top 6 weighted subspaces in $C$ for a
  specific column in $P$. Each subspace in $C$ is two-dimensional and we show
  the unwhitened image domain weights for both subspaces in A) and B). The
  corresponding image domain amplitudes and phases for the subspaces are shown
  in C) and D). There is clear grouping of subspaces with similar positions,
  orientations, and scales.}
\label{fig:P}
\end{figure}

Here we examine the structure represented in the model parameters $R$, $Q$,
$P$, and $C$ after training the mpkRBM on natural images. The subspace filters
in the $C$ learn localized oriented band-pass filters roughly in quadrature,
see Fig.~\ref{fig:C_mpkRBM}. We have observed that the filters in the matrix
$C$ appear to learn more textured patterns than those in the mcRBM, but a more
rigorous analysis is needed to verify such an observation. The weights in the
matrix $P$ adapt to group subspaces with similar spatial position and spatial
frequency. See Fig.~\ref{fig:P} for a depiction of the image filters with the
highest weights to each hidden unit $h^p$. The values in the learned matrix
$W$ are similar to those learned by the mcRBM and are shown in
Fig.~\ref{fig:W}.

\begin{figure}
\[
\includegraphics[width=\linewidth]{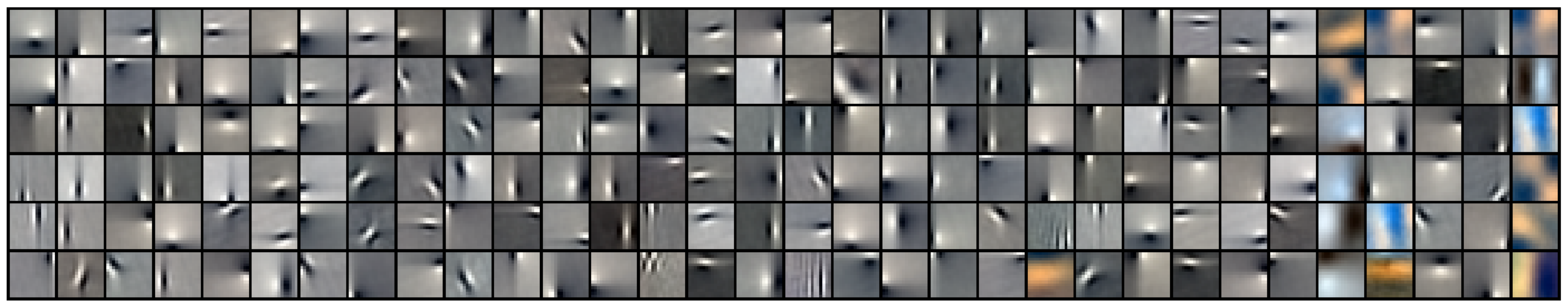}
\]
\caption{Learned Phase Projections, $Q$: The first 32 (out of 256) columns in
  $Q$ are shown. The entries in each column of $Q$ weight the cosine or sine
  of each subspace. Because the cosine and sine correspond to specific vectors
  in the tensor $C$, we show the image domain projection of these vectors that
  take the highest weight in the column of $Q$. In this figure, the 6 image
  domain projections with the highest magnitude weights are shown in the rows
  for different columns of $Q$ (each is shown in a different column of the
  figure).}
\label{fig:Q}
\end{figure}

\begin{figure}
\[
\includegraphics[width=\linewidth]{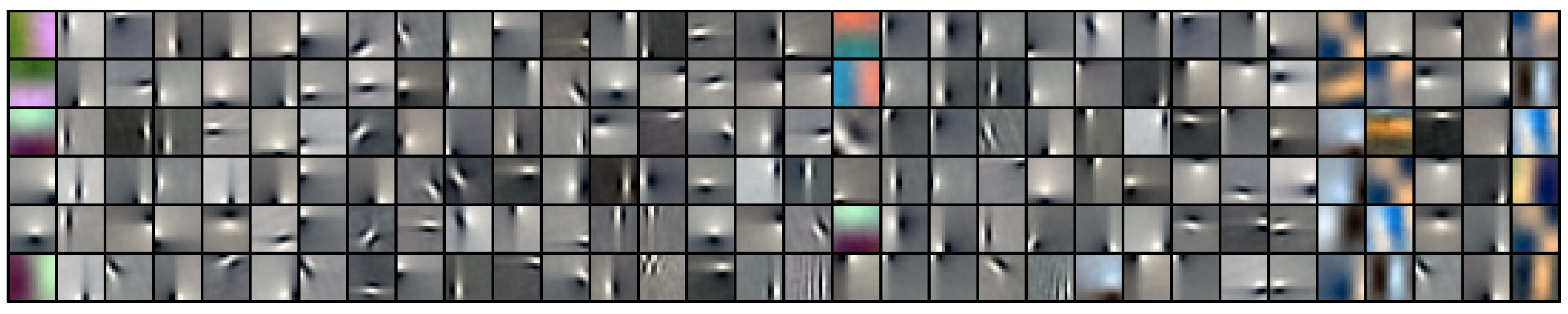}
\]
\caption{Learned Phase Coupling, $R$: The first 32 (out of 256) columns in
  $R$. Each column in $R$ produces a different coupling matrix (see
  Eq.~\ref{eq:K_coupling}). The values in this matrix indicate phase coupling
  between pairs of subspaces. Therefore, for the matrix $K$ produced by a
  specific column of $R$, we find the couplings with the highest
  magnitude. Given these sorted pairs, we take the unique top 6
  entries. Finally, we plot the image domain filters corresponding to these 6
  entries (shown in the rows). As in Fig.~\ref{fig:Q} these entries can be
  mapped to vectors in the tensor $C$ and thus plotted in the image
  domain. Different columns of $R$ are shown in different columns of the
  figure.}
\label{fig:R}
\end{figure}

The learned $R$ and $Q$ weights are harder to visualize as they express
dependencies in a layer removed from the image domain. However, we can view
the subspaces that are weighted highest by each column of $Q$. For each column
in Fig.~\ref{fig:Q} we depict the image domain filters ($C$) that are weighted
highest by the corresponding column in $Q$. We similarly show the image domain
filters that are weighted highest by each column in $R$ in Fig.~\ref{fig:R}.

\section{Discussion}
The mpRBM and mpkRBM suggest a number of interesting future directions. For
example, it should be possible to learn the dimensionality of the subspaces by
introducing a weighting matrix in the $L$ dimensional space of the tensor
$C$. However, it is not clear how to define an appropriate angle within these
subspaces for the phase-coupling factors. Although, it would be reasonable to
learn a separate set of factors, $C$, for the $h^p$ units and the $h^k$ units,
thus freeing the constraint of $L=2$ for the $h^p$ units. It may also be
possible to extend the mpRBM to a nested form of $L_p$ distributions suggested
in~\cite{sinz2010}. It is also worth exploring the behavior of the
phase-coupling hidden units as the number of hidden units is varied. As we had
little prior expectation as to the structure of the learned $R$ and $Q$
matrices, the choice of 100 $h^k$ hidden units was rather arbitrary.

\section{Conclusions}
In this paper we have introduced two new factorized Boltzmann machines: the
mpRBM and the mpkRBM, which each extend the factorized third-order Boltzmann
machine (the mcRBM) of Ranzato and Hinton~\cite{ranzato2010}. The form of
these additional hidden unit factors are motivated by image models of subspace
structure~\cite{Koster2007} and phase alignments due to edges in natural
images~\cite{kovesi2000}. Focusing on the mpkRBM, we have shown that such a
model learns phase structure in natural images.

\section{Acknowledgments}
We would like to thank Marc'Aurelio Ranzato for helpful comments and
discussions. We would also like to thank Bruno A. Olshausen for his
contributions to early drafts of this document. This work has been supported
by NSF grant IIS-0917342 (KK) and NSF grant IIS-0705939 to Bruno A. Olshausen.

\vfill

\pagebreak
\renewcommand\refname{\normalsize References} 


\begin{thebibliography}{10}

\bibitem{ranzato2010}
M.~Ranzato and G.~Hinton.
\newblock Modeling pixel means and covariances using factorized third-order
  boltzmann machines.
\newblock In {\em Proc. of Computer Vision and Pattern Recognition Conference
  (CVPR 2010)}. 2010.

\bibitem{karklin2008emergence}
Y.~Karklin and M.S. Lewicki.
\newblock {Emergence of complex cell properties by learning to generalize in
  natural scenes}.
\newblock {\em Nature}, 457(7225):83--86, 2008.

\bibitem{Lyu2009}
S.~Lyu and E.P. Simoncelli.
\newblock Modeling multiscale subbands of photographic images with fields of
  {Gaussian} scale mixtures.
\newblock {\em IEEE Trans. Patt. Analysis and Machine Intelligence},
  31(4):693--706, April 2009.

\bibitem{Lyu2009a}
S.~Lyu and E.P. Simoncelli.
\newblock {Nonlinear Extraction of Independent Components of Natural Images
  Using Radial Gaussianization}.
\newblock {\em Neural Computation}, 21(6):1485--1519, 2009.

\bibitem{Sinz2008}
F.~Sinz and M.~Bethge.
\newblock {The Conjoint Effect of Divisive Normalization and Orientation
  Selectivity on Redundancy Reduction in Natural Images}.
\newblock In {\em Frontiers in Computational Neuroscience. Conference Abstract:
  Bernstein Symposium 2008}, 2008.

\bibitem{Koster2007}
U.~Koster and A.~Hyvarinen.
\newblock {A two-layer ICA-like model estimated by score matching}.
\newblock {\em Lecture Notes in Computer Science}, 4669:798, 2007.

\bibitem{CadieuNIPS2009}
C.~F. Cadieu and B.~A. Olshausen.
\newblock Learning transformational invariants from natural movies.
\newblock In D.~Koller, D.~Schuurmans, Y.~Bengio, and L.~Bottou, editors, {\em
  Advances in Neural Information Processing Systems 21}, pages 209--216. MIT
  Press, 2009.

\bibitem{kovesi2000}
P.~Kovesi.
\newblock Phase congruency: A low-level image invariant.
\newblock In {\em Psychological Research Psychologische Forschung}, volume~64,
  pages 136--148. Springer-Verlag., 2000.

\bibitem{cadieu2010}
C.~F. Cadieu and K.~Koepsell.
\newblock Phase coupling estimation from multivariate phase statistics.
\newblock {\em Neural Computation}, 22(12):3107--3126, 2010.

\bibitem{Karklin2005}
Y.~Karklin and M.S. Lewicki.
\newblock A hierarchical bayesian model for learning nonlinear statistical
  regularities in nonstationary natural signals.
\newblock {\em Neural Computation}, 17(2):397--423, 2005.

\bibitem{Schwartz2001}
O.~Schwartz and E.P. Simoncelli.
\newblock {Natural signal statistics and sensory gain control}.
\newblock {\em Nature neuroscience}, 4(8):819--825, 2001.

\bibitem{Zetzsche1999}
C.~Zetzsche, G.~Krieger, and B.~Wegmann.
\newblock The atoms of vision: Cartesian or polar?
\newblock {\em Journal of the Optical Society of America A}, 16(7):1554--1565,
  1999.

\bibitem{Hinton2002}
G.E. Hinton.
\newblock {Training products of experts by minimizing contrastive divergence}.
\newblock {\em Neural Computation}, 14(8):1771--1800, 2002.

\bibitem{sinz2010}
F.~Sinz, E.~P. Simoncelli, and M.~Bethge.
\newblock Hierarchical modeling of local image features through lp-nested
  symmetric distributions.
\newblock In {\em Adv. Neural Information Processing Systems 22}, volume~22,
  pages 1696--1704. May 2010.

\bibitem{hinton2006}
G.~E. Hinton, S.~Osindero, M.~Welling, and Y.~Teh.
\newblock Unsupervised discovery of non-linear structure using contrastive
  backpropagation.
\newblock {\em Cognitive Science}, 30(4):725--731, 2006.

\bibitem{Neal1993}
R.M. Neal.
\newblock Probabilistic inference using markov chain monte carlo methods.
\newblock Technical Report CRG-TR-93-1, Dept. of Computer Science, University
  of Toronto, 1993.

\end{thebibliography}
\end{document}